\documentclass[letterpaper, 10 pt, conference]{ieeeconf}  

\IEEEoverridecommandlockouts                              

\usepackage{graphicx}
\usepackage{amsmath,amssymb,amsfonts}
\usepackage{algorithmic}
\usepackage{algorithm}
\usepackage{xcolor}
\usepackage{animate}
\usepackage{subfig}
\usepackage{tabularx,booktabs}
\usepackage{url}

\title{\LARGE \bf
Learning to SLAM on the Fly in Unknown Environments: A Continual Learning Approach for Drones in Visually Ambiguous Scenes
}

\author{Ali Safa$^{1,4}$, Tim Verbelen$^{2,4}$, Ilja Ocket$^{4}$, André Bourdoux$^{4}$, Hichem Sahli$^{3,4}$, \\
Francky Catthoor$^{1,4}$, Georges G.E. Gielen$^{1,4}$
\thanks{This research has received funding from the Flemish Government (AI Research Program) and the European Union's ECSEL Joint Undertaking under grant agreement n° 826655 - project TEMPO.}
\thanks{$^{1}$ Faculty of Electrical Engineering (ESAT) KU Leuven, 3001, Belgium}%
\thanks{$^{2}$ IDLab, Ghent University, B-9052 Gent, Belgium}%
\thanks{$^{3}$ ETRO, VUB, 1050 Brussels, Belgium}
\thanks{$^{4}$ imec, Kapeldreef 75, 3001, Leuven, Belgium
        {\tt\small \{Ali.Safa, Tim.Verbelen, Ilja.Ocket, Andre.Bourdoux, Hichem.Sahli, Francky.Catthoor\}@imec.be},
        {\tt\small Georges.Gielen@kuleuven.be}}%
}

\begin{document}

\maketitle
\thispagestyle{empty}
\pagestyle{empty}

\begin{abstract}

Learning to safely navigate in unknown environments is an important task for autonomous drones used in surveillance and rescue operations. In recent years, a number of learning-based Simultaneous Localisation and Mapping (SLAM) systems relying on deep neural networks (DNNs) have been proposed for applications where conventional feature descriptors do not perform well. However, such learning-based SLAM systems rely on DNN feature encoders trained offline in typical deep learning settings. This makes them less suited for drones deployed in environments unseen during training, where continual adaptation is paramount. In this paper, we present a new method for learning to SLAM on the fly in unknown environments, by modulating a low-complexity Dictionary Learning and Sparse Coding (DLSC) pipeline with a newly proposed Quadratic Bayesian Surprise (QBS) factor. We experimentally validate our approach with data collected by a drone in a challenging warehouse scenario, where the high number of ambiguous scenes makes visual disambiguation hard. 

\end{abstract}

\section*{Multimedia material}
A video showing our continual learning SLAM pipeline is provided at \textit{\url{http://tinyurl.com/ycyc5upc}}.
\section{Introduction}
Simultaneous Localisation and Mapping (SLAM) is a fundamental task for autonomous agents such as surveillance and rescue drones \cite{uavforindoorfire}. State-of-the-art SLAM systems either rely on handcrafted feature descriptors \cite{ratslam, orbslam2} or on learning-based representations \cite{contSLAM, latentslam} for template matching and loop closure detection. Compared to handcrafted descriptors, learning-based systems perform better in environments with less features or with high amounts of visual ambiguities (i.e., different scenes that look similar) \cite{latentslam, alias}. 

Still, most learning-based SLAM systems use traditional deep neural networks (DNNs), where a dataset of the target environment must be available \textit{a priori} for an offline DNN training phase \cite{latentslam}. Hence, those learning-based systems do not always generalize well to environments unseen during training \cite{contSLAM}. This has recently motivated the study of continual learning (CL) for SLAM \cite{contSLAM}.

In the past years, CL has gained much attention and has been mostly investigated in the context of deep learning,
\begin{figure}[H]
\centering
    \includegraphics[scale = 0.47]{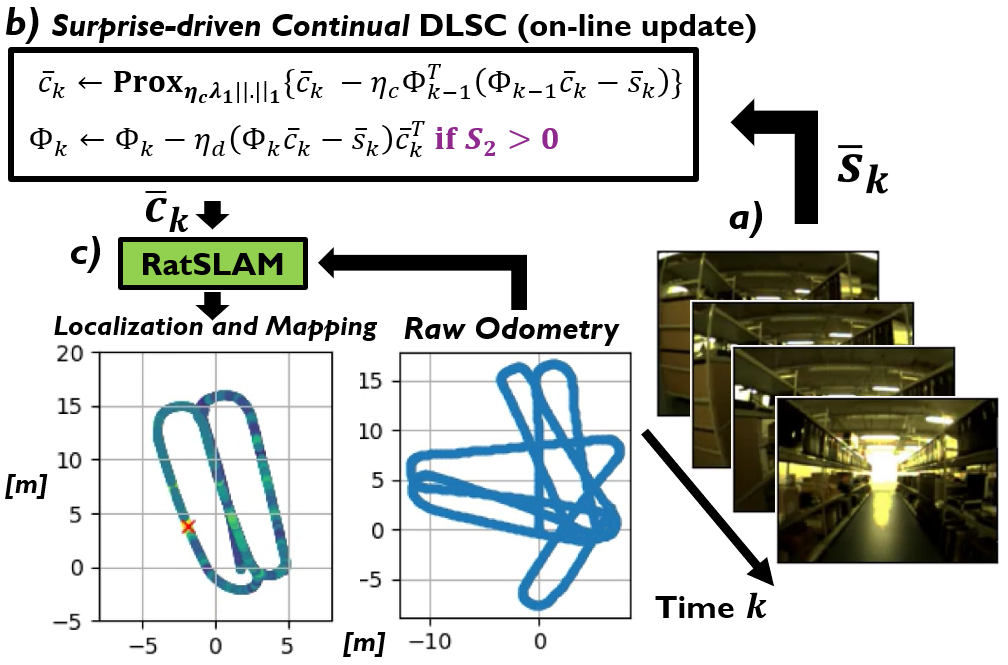}
    \caption{\textit{\textbf{Proposed Continual Dictionary Learning approach} for performing SLAM in unknown environments (without any offline pre-training). a) The consecutive images captured by a flying drone are fed in their natural order to a surprise-driven dictionary learning and Sparse Coding (DLSC) pipeline. b) The DLSC continuously infers a latent code $\bar{c}_k$ corresponding to the current observation $\bar{s}_k$. At the same time, DLSC continuously learns a dictionary $\Phi$ when the proposed Quadratic Bayesian Surprise (QBS) factor $\mathcal{S}_2$ is positive. c) the latent codes $\bar{c}_k$ are fed to a RatSLAM back-end \cite{ratslam} to perform loop closure detection through template matching \cite{latentslam}.}}
    \label{entryconcept}
\end{figure}
\begin{figure}[htbp]
\centering
    \includegraphics[scale = 0.66]{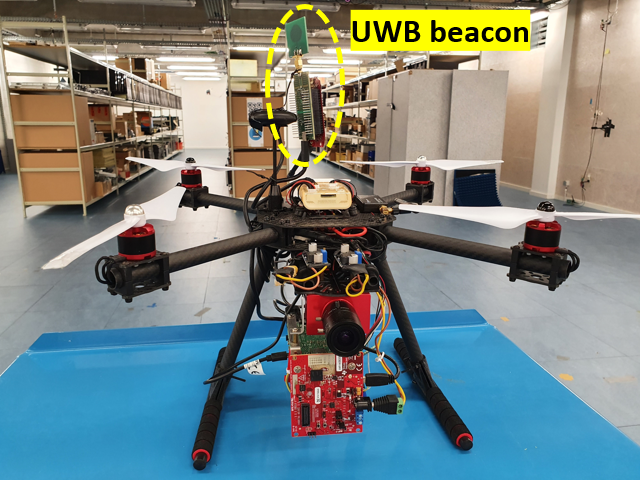}
    \caption{\textit{\textbf{Our multi-sensor drone} used for data acquisition (RGB camera data and odometry solely used in this work). A view of the warehouse environment (described in \cite{indooruwb}) where data was acquired is also shown. The warehouse is equipped with an Ultra Wide Band (UWB) localisation system for ground truth labelling of the drone's trajectory \cite{indooruwb}. }}
    \label{dronesetup}
\end{figure}
\noindent
with the aim of training deep neural networks on streams of non-shuffled data which cannot be assumed independent and identically distributed (i.i.d.). This poses significant challenges during training
, severely jeopardizing performance \cite{nonstatdatastream}. Most notably, research in CL has been mainly devoted to classification tasks, using non-i.i.d versions of popular datasets such as MNIST, CIFAR10 and ImageNet \cite{nonstatdatastream}.

More recently, Simultaneous Localisation and Mapping (SLAM) has also been proposed as an interesting application for CL \cite{contSLAM}. Indeed, when adopting a learning-based SLAM system, a robot trained offline on a dataset acquired beforehand could perform extremely poorly in environments that were not captured by the dataset \cite{contSLAM}, leading to unreliable and unsafe navigation. Of course, this situation is paramount in the context of search and rescue, which requires robust navigation in extreme and harsh environments \cite{uavforindoorfire}.




Therefore, to enable the deployment of learning-based SLAM systems in unseen environments, this paper proposes continual Dictionary Learning and Sparse Coding (DLSC), a fast and robust CL system for SLAM, and a low-complexity yet high-performance alternative to DNN approaches. The contributions of this paper are:
 \begin{enumerate}
      \item We propose a new \textit{Quadratic Bayesian Surprise} (QBS) for modulating DLSC, enabling continual learning.
      \item We experimentally demonstrate our continual learning DLSC-QBS method for performing SLAM in the important context of environments not captured by any dataset beforehand, without any offline pre-training.
  \end{enumerate}


This paper is organized as follows. First, related work is reviewed in Section \ref{related}. Next, necessary background is given in Section \ref{preli}. Our methods are provided in Section \ref{props} and results are presented in Section \ref{expres}. Finally, conclusions are provided in Section \ref{concs}.

\section{Related work}
\label{related}
A number of SLAM architectures have been proposed in the past decades, exploring the use of handcrafted features and learned representations \cite{ratslam, orbslam2, contSLAM, latentslam}. 

Related to this work, the popular RatSLAM architecture has been proposed as a bio-inspired system modelled following the navigational processes taking place in the rat hippocampus \cite{ratslam}. In contrast to our proposed system, RatSLAM does not use learned representations, but opts for a simple template matching based on raw RGB data in order to detect loop closures. This simple template matching was later shown to be inefficient in environments with lots of ambiguous views (as in the case of our work) \cite{latentslam,alias}. 

More recently, the LatentSLAM architecture has been proposed in \cite{latentslam}, outperforming RatSLAM through the use of a DNN encoder outputting latent codes for template matching, trained offline as a variational autoencoder. In contrast to LatentSLAM, our CL system does not require any offline training phase, enabling its deployment in unseen environments (not captured by a dataset available \textit{a priori}).

In this work, we also deviate from most learning-based SLAM systems by using a lower-complexity, unsupervised DLSC learning pipeline \cite{ksvd} instead of conventional DNNs \cite{latentslam, contSLAM}. Similar to LatentSLAM \cite{latentslam}, we feed the feature descriptors produced by DLSC to the loop closure detection and map correction back-end of the RatSLAM system \cite{ratslam}.

In Section \ref{expres}, we benchmark our system against both RatSLAM and LatentSLAM, and against the use of hand-crafted ORB feature matching \cite{orbfeature}, extensively used in state-of-the-art SLAM systems \cite{orbslam2}.

\section{Background theory}
\label{preli}
\subsection{Joint Dictionary Learning and Sparse Coding (DLSC)}
\label{DLSC}
DLSC is concerned with the problem of jointly learning a dictionary $\Phi$ of size $N\times M$ and inferring output codes $\bar{c}_k$ of size $M$ from a stream of input data $\bar{s}_k$ of size $N$ \cite{ksvd}: 
\begin{equation}
    C, \Phi = \arg \min_{C, \Phi} \sum_{k=1}^{K} \frac{1}{2}||\Phi \bar{c}_k - \bar{s}_k ||_2^2 + \lambda_1 ||\bar{c}_k||_1 
    \label{problem}
\end{equation}
where $C = [\bar{c}_1, \bar{c}_2, \hdots , \bar{c}_K]$ contains all output vectors $\bar{c}_k$. The first term in (\ref{problem}) seeks to minimize the re-projection error between the output sparse code $\bar{c}_k$ and the input $\bar{s}_k$ through the dictionary $\Phi$. The second term provides $l_1$ regularization, controlling the sparsity of the codes (as in LASSO \cite{ksvd}).

Since our goal is to use the codes $\bar{c}_k$ as global descriptors for performing template matching in SLAM systems, we consider an under-complete dictionary $\Phi$ with $M \ll N$ to infer a lower-dimensional latent code for each input $\bar{s}_k$ \cite{latentslam}.  

Conventionally, it is assumed that the input sequence $\bar{s}_k$ originates from a shuffled dataset (i.e., the realizations $\bar{s}_k$ are assumed i.i.d) and the DLSC problem in (\ref{problem}) is classically solved by alternating between a stochastic gradient descent (SGD) step for the learning of $\Phi$ and a proximal descent step for the inference of $\bar{c}_k$ \cite{ksvd, daubechies}. The proximal operator to the $l_1$ norm being defined as \cite{daubechies}:
\begin{equation}
   \textbf{Prox}_{\eta_c \lambda_1 ||.||_1}(\bar{x}) = \max (0, \bar{x} - \eta_c \lambda_1) + \min (0, \bar{x} + \eta_c \lambda_1) 
   \label{proxop}
\end{equation}

The DLSC problem (\ref{problem}) is therefore a good starting point for setting up an unsupervised encoder that can jointly infer latent codes while learning features from the environment. Still, conventional DLSC solving via alternating descent \cite{ksvd} (summarized above) does not cover the continual learning case since it requires a shuffled dataset. This issue will be addressed in Section \ref{props} with our proposed Algorithm \ref{pursuitalg}.

\subsection{Continual Learning Challenges}
\label{challenge}
As already stated in Section \ref{DLSC}, the input sequence $\bar{s}_k$ is generally assumed to be i.i.d following the conventional training procedure where a dataset is assumed to be available \textit{a priori} and shuffled before being fed to the optimisation procedure of Algorithm \ref{pursuitalg}. Violating this i.i.d assumption \textit{as in this work}, by feeding video sequences in their natural order, causes significant problems during learning since the stochastic gradients are not representative of the full loss \cite{nonstatdatastream}:
\begin{equation}
    \mathbb{E}[\frac{\partial l(\bar{s}_k, \Phi)}{\partial \Phi}] \not\to \frac{\partial \frac{1}{K}\sum_{k=1}^{K} l(\bar{s}_k, \Phi)}{\partial \Phi}
    \label{reasonwhy}
\end{equation}
where $\mathbb{E}$ denotes the expected value, $K$ is the total number of data points and $l(\bar{s}_k,\Phi) = \frac{1}{2}||\Phi \bar{c}_k - \bar{s}_k ||_2^2 + \lambda_1 ||\bar{c}_k||_1 $ is the loss associated to the data point $\bar{s}_k$ (used by SGD).

In this work, our goal is to enable continual DLSC in non-i.i.d data streams such as videos where the consecutive image frames are highly correlated \cite{nonstatdatastream} ($k$ is the time index):
\begin{equation}
    \bar{s}_{k+1} \approx \bar{s}_{k}
    \label{nonstat}
\end{equation}

\begin{figure}[htbp]
\centering
    \includegraphics[scale = 0.38]{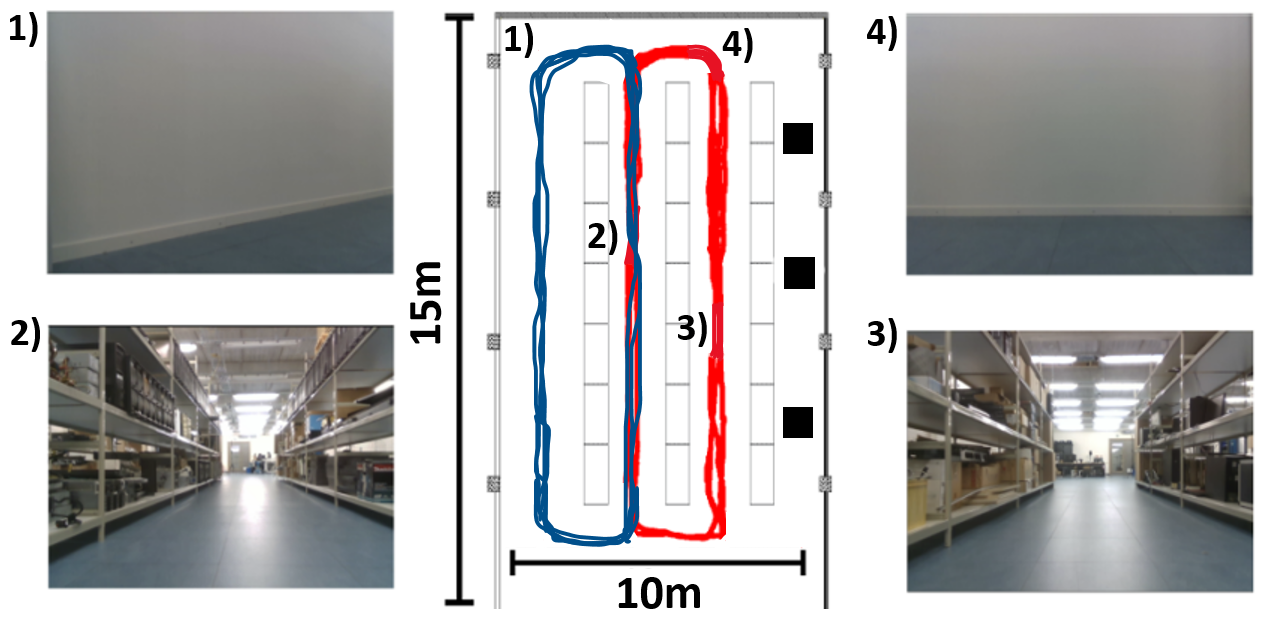}
    \caption{\textit{\textbf{Flight data used in this work.} Drone location is hard to disambiguate with RGB (e.g., view 1 vs. 4, 2 vs. 3). Data has been acquired during three different flight sequences: the blue path, the red path and a combination of both. }}
    \label{aliasing}
\end{figure}

Therefore, our goal is to continuously learn $\Phi$ and infer the latent codes $\bar{c}_k$ from a video stream without any pre-training and without re-initializing $\bar{c}$ between each frames (see Fig. \ref{entryconcept}). As a challenging, real-world use case, we use the drone setup in Fig. \ref{dronesetup} to illustrate the applicability of our method for learning to SLAM on the fly in a highly ambiguous warehouse environment (see Fig. \ref{aliasing}). 

This high redundancy (\ref{nonstat}) poses a significant challenge for both the continual learning aspect \cite{nonstatdatastream} due to the SGD bias (\ref{reasonwhy}), and to the loop closure detection, due to the high levels of aliasing in the environment, as shown in \cite{alias} (see Fig. \ref{aliasing}). 


\section{Proposed Method}
\label{props}
\begin{figure*}[!t]
\centering
\captionsetup{justification=centering}
    \includegraphics[scale = 0.555]{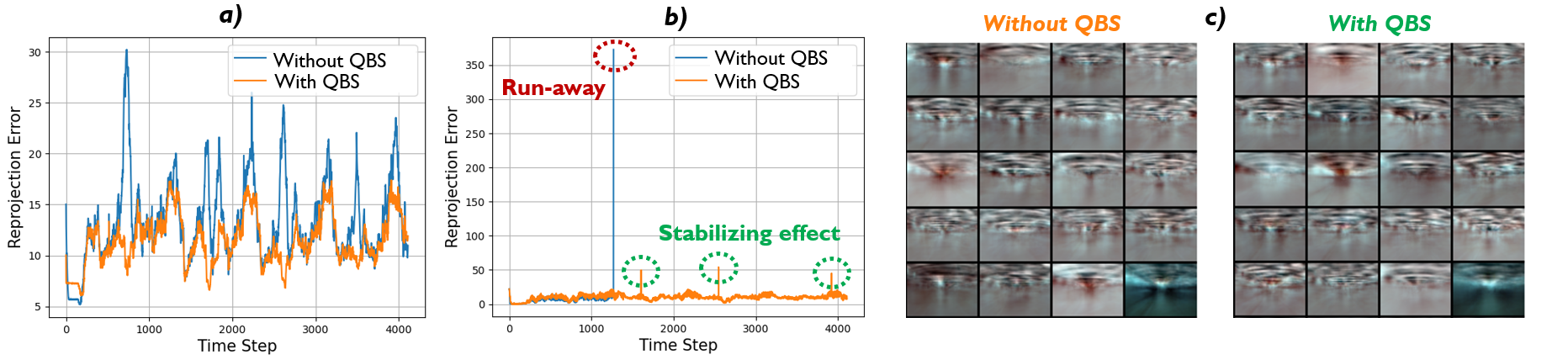}
    \caption{\textit{\textbf{Effect of the QBS on DLSC performance and stability.} a) Reprojection error obtained after continual dictionary learning on an RGB video sequence captured with our drone setup (images of size $N=346 \times 260$). The sequence contains long episodes of ambiguous views (see Fig. \ref{aliasing}) and is therefore challenging due to high over-fitting risks. The use of the QBS factor leads to a lower loss during sequence replay, indicating a better learning performance (parameters: $\eta_c = 5\times 10^{-3}$, $\lambda_1 = 0.5$, $\eta_d = 2\times 10^{-3}$, $N_{c} = 10$, $N_d = 1$, $M = 64$ in Algorithm \ref{pursuitalg}) b) Without the use of the surprise factor, severe run-away behaviors can happen during the continual learning process (parameters: $\eta_c = 10^{-2}$,  $\lambda_1 = 0.25$, $\eta_d = 4\times 10^{-3}$, $N_{c} = 10$, $N_d = 1$, $M = 64$ in Algorithm \ref{pursuitalg}). c) Subset of the dictionaries learned during the CL process.  
    }}
    \label{bigfig}
\end{figure*}
In order to cast (\ref{problem}) into the continual learning case, we use the DLSC reprojection error $||\Phi \bar{c}_k - \bar{s}_k ||_2^2$ in (\ref{problem}) to reduce the coding problem (inference of $\bar{c}_k$ by LASSO proximal descent) into a 2-class likelihood test. Indeed, the DLSC reprojection is linked to the Gaussian likelihood $p(\bar{s}_k|\Phi)$:
\begin{equation}
    -\log p(\bar{s}_k|\Phi) \sim ||\Phi \bar{c}_k - \bar{s}_k ||_2^2
\end{equation}
This likelihood can be used to perform an inlier-outlier test:
\begin{equation}
    \begin{cases}
    \textbf{if } p(\bar{s}_k|\Phi) \geq \theta \textbf{ Then } H_0
    \\
    \textbf{if } p(\bar{s}_k|\Phi) < \theta \textbf{ Then } H_1
    \end{cases}
    \label{conceptprob}
\end{equation}

The test (\ref{conceptprob}) can be seen as a binary classification problem inferring whether the current input $\bar{s}_k$ corresponds to the model learned in the dictionary $\Phi$ ($H_0$) or not ($H_1$). This classification is done by testing $p(\bar{s}_k|\Phi)$ against a threshold $\theta$. Therefore, we identify two dual problems: \textit{A)} the original problem that we are seeking to solve (i.e., learning $\Phi$ in a continual manner in order to infer $\bar{c}$) and \textit{B)} its associated dual problem as learning $\Phi$ in a {continual manner in order to classify $\bar{s}$. Hence, by casting the dual problem \textit{B)} to the continual learning case, we implicitly cast the original problem \textit{A)} to the continual learning case as well. 

In order to transpose \textit{B)} to the continual case, we must provide robustness to the severe data imbalance caused by the non-stationary stream $\bar{s}_k$ (due to the fact that $\bar{s}_k$ does not originate from a shuffled dataset where realisations are assumed to be i.i.d, see Section \ref{challenge}). Therefore, we propose to solve the binary classification problem \textit{B)} using the well-known Support Vector Classification (SVC) framework \cite{hinge}: 
\begin{multline}
    \Phi_{k^*} = \arg \min_{\Phi} \sum_{j=1}^{k^*} \xi_j \hspace{20pt} \textbf{subject to:}
    \\
    \begin{cases}
    ||\Phi \bar{c}_j - \bar{s}_j ||_2^2 \leq \log \theta + \xi_j \textbf{ if } \bar{s}_j \in H_0
    \\
    ||\Phi \bar{c}_j - \bar{s}_j ||_2^2 \geq \log \theta - \xi_j \textbf{ if } \bar{s}_j \in H_1
    \end{cases}
    \label{dualproblem}
\end{multline}
where $k^*$ is the index of the current time step in the data stream, $\theta$ is the threshold used in (\ref{conceptprob}), and $\xi_j$ are the slack variables defining the margin of the separation hyperplane.

The SVC formulation in (\ref{dualproblem}) is indeed a natural choice when dealing with class imbalance since in an SVC, all data points are not contributing to the synthesis of the decision hyperplane \cite{hinge} (the contribution is only due to a few support vectors). This prevents SVC over-fitting on the redundant data found in our heavily imbalanced video streams.

We can then cast (\ref{dualproblem}) into its \textit{hinge loss} formulation \cite{hinge}:
\begin{multline}
    \Phi_{k^*} = \arg \min_{\Phi} \sum_{j \in H_0} \max (0, ||\Phi \bar{c}_j - \bar{s}_j   ||_2^2 - \log \theta)
    \\
    + \sum_{j \in H_1} \max (0, \log \theta - ||\Phi \bar{c}_j - \bar{s}_j   ||_2^2)
    \label{toconsider}
\end{multline}
where the hinge loss $\max (0, x)$ leads to the sparsity property of SVCs, which in turn leads to a large number of input vectors $\bar{s}_k$ not playing any role in the synthesis of the SVC hyperplane. This provides the robustness to data redundancy.

In our particular case, we seek to continuously maintain a dictionary $\Phi$ such that all incoming data points are well modelled by this dictionary. Therefore, the generic 2-class formulation in (\ref{toconsider}) reduces to (\ref{oneclassprob}) in our case since all data points must be associated to the $H_0$ class to be inliers \cite{oneclass}:
\begin{equation}
    \Phi_{k^*} = \arg \min_{\Phi} \sum_{j = 1}^{k^*} \max (0, ||\Phi \bar{c}_j - \bar{s}_j   ||_2^2 - \log \theta)
    \label{oneclassprob}
\end{equation}

Using (\ref{oneclassprob}) in a probabilistic context, we can define the \textit{Bayesian surprise factor} $\mathcal{S}_1$ associated with the incoming data point $\bar{s}_{k^*+1}$ as the posterior to prior ratio \cite{bayessurprise}:
\begin{multline}
\hspace{-9pt} \mathcal{S}_1(k^* + 1) = -\log \frac{p(\Phi_{k^*+1}|\bar{s}_{k^*+1})}{p(\Phi_{k^*})} 
\sim - \log p(\bar{s}_{k^*+1}|\Phi_{k^*}) 
\\
\propto \max (0, ||\Phi_{k^*} \bar{c}_{k^*+1} - \bar{s}_{k^*+1}   ||_2^2 - \log \theta)
\label{surprise1}
\end{multline}
where $\mathcal{S}_1$ is proportional to the negative log-likelihood via the Bayes theorem. In addition, the formulation in (\ref{surprise1}) assumes that the negative log-likelihood $- \log p(\bar{s}_{k^*+1}|\Phi_{k^*}) $ is proportional to the hinge loss energy term in (\ref{toconsider}) under the SVC formulation (\ref{oneclassprob}). The surprise factor $\mathcal{S}_1$ is null when the hinge loss associated with the incoming data point $\bar{s}_{k^*+1}$ is zero, denoting that the data point is already well modelled by the dictionary $\Phi$ and hence, does not lead to any surprise.  

Still, using $\mathcal{S}_1$ in (\ref{surprise1}) is inconvenient since it requires the knowledge of the likelihood threshold $\theta$. This can be an even greater problem since an optimal choice of $\theta$ could slowly vary over time due to changes in the environment. Under the generic assumption of a high-enough camera frame rate (\ref{nonstat}), we now show that having an explicit knowledge of $\theta$ can be eliminated by introducing a second-order surprise factor $\mathcal{S}_2$:
\begin{multline}
    \mathcal{S}_2(k^*+1) \equiv \mathcal{S}_1(k^*+1) - \mathcal{S}_1(k^*) 
    \\
    = -\log \frac{p(\Phi_{k^*+1}|\bar{s}_{k^*+1})}{p(\Phi_{k^*})} + \log \frac{p(\Phi_{k^*}|\bar{s}_{k^*})}{p(\Phi_{k^*-1})}
    \\
    \sim \frac{\delta^2 -\log p(\Phi_{k}|\bar{s}_{k}) }{\delta k^2}|_{k^*+1}
    \label{secondordersurpise}
\end{multline}
which is related to the discrete-time second derivative ($\frac{\delta^2}{\delta k^2}$) of the posterior density evolution through time. We thus refer to $\mathcal{S}_2$ as a \textit{Quadratic Bayesian Surprise} (QBS) factor.

Intuitively, $\mathcal{S}_2$ can be understood as the curvature of the posterior density evolution through time, indicating how much a newly received input vector $\bar{s}_{k^*+1}$ will accelerate or decelerate the change in posterior $p(\Phi_{k}|\bar{s}_{k})$. 

Dropping proportionality constants and using (\ref{surprise1}) in (\ref{secondordersurpise}):
\begin{multline}
    \mathcal{S}_2 \equiv \max (0, \overbrace{||\Phi_{k^*} \bar{c}_{k^*+1} - \bar{s}_{k^*+1}   ||_2^2}^{e_{k^*+1}} - \log \theta)
    \\
    - \max (0, \overbrace{||\Phi_{k^*-1} \bar{c}_{k^*} - \bar{s}_{k^*}   ||_2^2}^{e_{k^*}} - \log \theta)
    \\
    =
    \begin{cases}
    0 \textbf{ if } e_{k^*+1} - \log \theta \leq 0, e_{k^*} - \log \theta \leq 0 \textbf{ a)}
    \\
    e_{k^*+1} - \log \theta \textbf{ if only } e_{k^*} - \log \theta \leq 0 \hspace{5pt} \textbf{ b)}
    \\
    \log \theta - e_{k^*} \textbf{ if only } e_{k^*+1} - \log \theta \leq 0 \hspace{6pt} \textbf{ c)}
    \\
    e_{k^*+1} - e_{k^*} \textbf{ else } \hspace{88pt} \textbf{ d)}
    \end{cases}
    \label{outcomes}
\end{multline}
where $e_{k^*+1}$ and $e_{k^*}$ are the current and past reprojection errors. Four possible outcomes \textit{a,b,c,d} are identified in (\ref{outcomes}). Under the high frame rate assumption (\ref{nonstat}), the probability of outcomes \textit{b} and \textit{c} in (\ref{outcomes}) will be very low since (\ref{nonstat}) induces that if $e_{k^*} - \log \theta \leq 0$ then $e_{k^*+1} - \log \theta \leq 0$ must likely be the case as well, and vice versa. 
In addition, outcomes \textit{a} and \textit{c} in (\ref{outcomes}) can be merged under the assumption (\ref{nonstat}):
\begin{multline}
    \textbf{a) } \textbf{ if } e_{k^*+1} - \log \theta \leq 0, e_{k^*} - \log \theta \leq 0
    \\
     \textbf{then } \mathcal{S}_2 = 0 \approx e_{k^*+1} - e_{k^*} \textbf{ using } \bar{s}_{k^*+1} \approx \bar{s}_{k^*}
\end{multline}
while still observing the conditions $e_{k^*+1} - \log \theta \leq 0, e_{k^*} - \log \theta \leq 0$ of \textit{a)} in (\ref{outcomes}). Therefore, the QBS in (\ref{secondordersurpise}) can simply be computed as follows:
\begin{equation}
    \mathcal{S}_2(k^*+1) \approx ||\Phi_{k^*} \bar{c}_{k^*+1} - \bar{s}_{k^*+1}||_2^2 - ||\Phi_{k^*-1} \bar{c}_{k^*} - \bar{s}_{k^*}   ||_2^2
    \label{quadraticnofil}
\end{equation}

In practice, we estimate the QBS (\ref{quadraticnofil}) by low-pass filtering $\mathcal{S}_2$ using a moving average of width $5$ to attenuate noise.

We can now use the QBS $\mathcal{S}_2$ to modulate SGD learning in an online manner using the following dictionary update procedure (with learning rate $\eta_d$) in Algorithm \ref{pursuitalg} (lines 6):
\begin{equation}
    \begin{cases}
    \textbf{if }\mathcal{S}_2(k^*+1)>0 \textbf{ , } \Phi_{k^*+1} \xleftarrow{} \Phi_{k^*} - \eta_d \frac{\partial e_{k^*+1} }{\partial \Phi}
    \\
    \textbf{else if } \mathcal{S}_2(k^*+1) \leq 0 \textbf{ , } \Phi_{k^*+1} \xleftarrow{} \Phi_{k^*}
    \end{cases}
    \label{continualrule}
\end{equation}

 \begin{algorithm}
 \caption{DLSC-QBS for SLAM (see Fig. \ref{entryconcept})}
 \label{pursuitalg}
 \begin{algorithmic}[1]
 \renewcommand{\algorithmicrequire}{\textbf{Input:}}
 \renewcommand{\algorithmicensure}{\textbf{Init.:}}
 \REQUIRE $\bar{s}_k \hspace{3pt} \forall k$: input vector stream, $\eta_c$: coding rate, $\eta_d$: learning rate, $\lambda_1$: regularization, $N_{c,d}$: number of iterations
 \ENSURE $\Phi_{ij} \xleftarrow{} \mathcal{N}(0,\sigma_w)$ (zero-mean normal distribution with std. deviation $\sigma_w \sim 0.01$), $\bar{c}_1 \xleftarrow{} 0$, $\mathcal{S}_2 \xleftarrow{} 1$
  \FOR {$k\in \{1,...,T_{end} \}$ (feed data sequence)}
  \FOR{$i = 1,...,N_{c}$ (local coding iterations)}
  \STATE $\bar{c}_k\xleftarrow{} \textbf{Prox}_{\eta_c \lambda_1 ||.||_1} \{\bar{c}_k- \eta_c \Phi^T(\Phi \bar{c}_k- \bar{s}_k) \}$ see (\ref{proxop})
  \ENDFOR
  \FOR{$i = 1,...,N_{d}$ (local learning iterations)} 
  \IF{$\mathcal{S}_2>0$}
  \STATE $\Phi_k \xleftarrow{} \Phi_{k-1} - \eta_d (\Phi \bar{c}_k- \bar{s}_k) \bar{c}^T_k$
  \ELSE
  \STATE $\Phi_k \xleftarrow{} \Phi_{k-1}$
  \ENDIF
  \ENDFOR
  \STATE Compute $\mathcal{S}_2$ following (\ref{quadraticnofil})
  \STATE $\textbf{RatSLAM}\{\bar{c}_k, \text{odometry}_k\}$ // Feed to SLAM
  \ENDFOR
 \end{algorithmic} 
 \end{algorithm}
 
Therefore, by ignoring the effect of redundant data associated to a negative QBS factor, we prevent the over-fitting of the dictionary learning to the local contexts in the video streams. In turn, this leads to the learning of a more diverse set of features in $\Phi$ and consequently, to lower reprojection errors during the replay of the input sequence (see Fig. \ref{bigfig}).  

In addition, we observed during our experiments a useful stability property induced by the use of the QBS in our continual dictionary learning scenario. Indeed, it is known that DLSC problems can suffer from instability \cite{stable1} and we have observed run-away behaviors as well for cases where the QBS was not being used. In contrast, we did not observe any instability issue when using our QBS-modulated learning rule (\ref{continualrule}) in Algorithm \ref{pursuitalg}, suggesting a regularizing effect (see Fig. \ref{bigfig} b, where using the QBS induces stability).

\section{Experimental Results}
\label{expres}
We evaluate the Mean Absolute Error (MAE) of our DLSC-QBS SLAM approach (see Fig. \ref{entryconcept}) against the ground truth localisation and mapping, acquired by UWB indoor positioning beacons \cite{indooruwb}. As DLSC parameters, we use $\eta_c = 5\times 10^{-3}$, $\eta_d = 1.4\times 10^{-3}$, $\lambda_1 = 0.2$, $N_{c} = 10$, $N_d = 1$, $M = 64$ in Algorithm \ref{pursuitalg}. These parameters were tuned empirically in order to optimise SLAM performance. 

Three different flight sequences are used for performance assessment, with increasing difficulty:
\begin{itemize}
    \item \textit{Fligh 1}: Flying between wall and shelves, which makes the continual learning and loop closure detection easier since the visited environment is more diverse and hence, easier to disambiguate (blue path in Fig. \ref{aliasing}). 
    \item \textit{Fligh 2}: Flying between shelves only, which leads to a larger amount of similar and ambiguous views compared to \textit{flight 1} (red path in Fig. \ref{aliasing}) 
    \item \textit{Fligh 3}: Combining \textit{flights 1 and 2} (both red and blue paths in Fig. \ref{aliasing} are visited by the drone).
\end{itemize}

Since the SLAM result and the ground truth do not share the same coordinate system \cite{ratslam}, the ground truth coordinates are projected to the SLAM domain via translation and rotation. The translation vector and rotation angle are found via grid search, by minimizing the localisation MAE. 

The localisation MAE is computed as the average error between the SLAM localisation $(x_{k}, y_{k})$ and the ground truth $(x_{k}^{\text{gt}}, y_{k}^{\text{gt}})$ on the flight sequence of length $T_{end}$:
\begin{equation}
    \text{MAE}_{\text{L}} = \frac{1}{T_{end}} \sum_{k=1}^{T_{end}} |x_{k} - x_k^{\text{gt}}| + |y_{k} - y_k^{\text{gt}}|
    \label{locerr}
\end{equation}

In addition, the mapping MAE is computed as the a posteriori deviation between the map obtained at the end of the SLAM process and all the ground truth points:
\begin{equation}
        \text{MAE}_{\text{M}} = \frac{1}{T_{end}} \sum_{k=1}^{T_{end}} \min_j \{ |x_{k} - x_j^{\text{gt}}| + |y_{k} - y_j^{\text{gt}}| \}
        \label{maperr}
\end{equation}

Table \ref{performanceloc} quantitatively compares the performance of our approach against \textit{i)} the original RatSLAM \cite{ratslam}; \textit{ii)} a VGG-11 pre-trained on ImageNet \cite{vgg} as feature encoder to RatSLAM; \textit{iii)} the LatentSLAM, trained specifically for our environment following \cite{latentslam}, on a set of 3998 frames acquired independently from flights 1-3; and \textit{iv)} ORB features for template matching \cite{orbfeature} following the \textit{Lowe's Ratio Test} \cite{lowe}. 

For the ORB features, the similarity between two frames $i,j$ is computed as $s_{ij} = 1- N_{m} / \max_{m}$ where $N_m$ is the number of matches and $\max_{m}$ the maximum number of possible matches. For VGG-11, LatentSLAM and DLSC-QBS, the similarity is computed as $s_{ij} = \frac{\bar{c}_i^T \bar{c}_j}{||\bar{c}_i ||_2 || \bar{c}_j||_2}$ with $\bar{c}_{i,j}$ denoting the feature descriptor obtained by each method.

The RatSLAM back-end integrates the raw odometry and detects loop closures by sampling latent codes from the data stream each $100$ms and by comparing the similarities $s_{ij}$ against a threshold $\mu$ tuned for minimum MAE (see Fig. \ref{entryconcept}).
\begin{table}[htbp]
\begin{tabularx}{0.47\textwidth}{@{}l*{2}{c}c@{}}
\toprule
Architecture  & Flight 1 &  Flight 2 &  Flight 3  \\ 
\midrule
RatSLAM \cite{ratslam}   &  0.72 $|$ 0.22        &   1.499 $|$ 0.248 &  2.714 $|$ 1.15 
\\ 
VGG-11 \cite{vgg}  &  3.95 $|$ 0.244        &  2.58 $|$ 1.15  &  1.775 $|$ 0.266 
\\ 
LatentSLAM \cite{latentslam}  &  0.54 $|$ 0.21        &  0.438 $|$ 0.239    &  \textbf{1.266} $|$ 0.274    
\\
ORB \cite{orbfeature}  &  \textbf{0.456} $|$ 0.165       &  \textbf{0.407} $|$ \textbf{0.215}    &  4.258 $|$ 1.595  
\\
\textbf{DLSC-QBS (Ours)}   &  0.588 $|$ \textbf{0.1572}      &  0.742 $|$ 0.329   &  1.596 $|$ \textbf{0.244} 
\\ 
\bottomrule
\end{tabularx}
\caption{\textit{\textbf{$\text{MAE}_{\text{L}}$ $|$ $\text{MAE}_{\text{M}}$}  (the lower the better). 
}
}
\label{performanceloc}
\end{table}


\begin{figure}[htbp]
\centering
    \includegraphics[scale = 0.41]{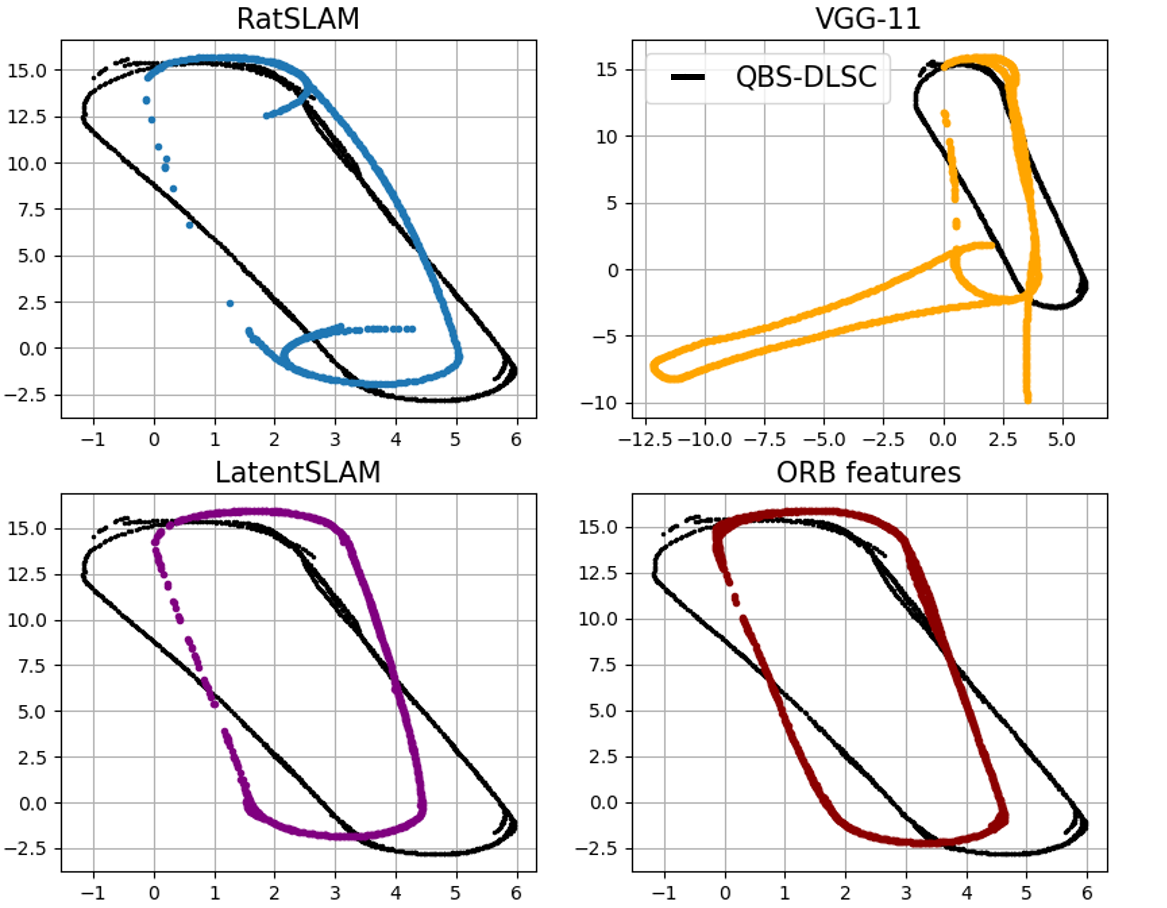}
    \caption{\textit{\textbf{Flight 1}, blue path in Fig. \ref{aliasing}. }}
    \label{seq1}
\end{figure}

\begin{figure}[htbp]
\centering
    \includegraphics[scale = 0.41]{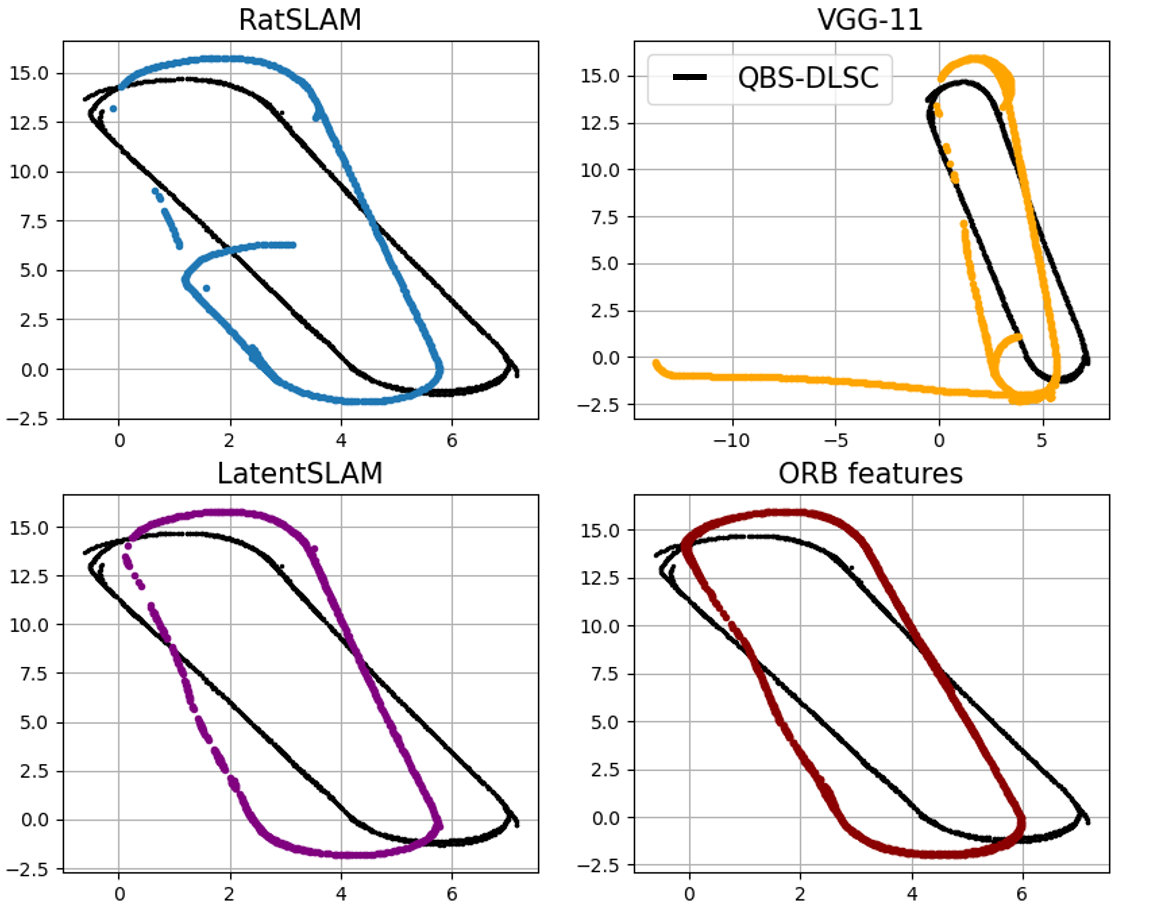}
    \caption{\textit{\textbf{Flight 2}, red path in Fig. \ref{aliasing}. }}
    \label{seq2}
\end{figure}

\begin{figure}[htbp]
\centering
    \includegraphics[scale = 0.41]{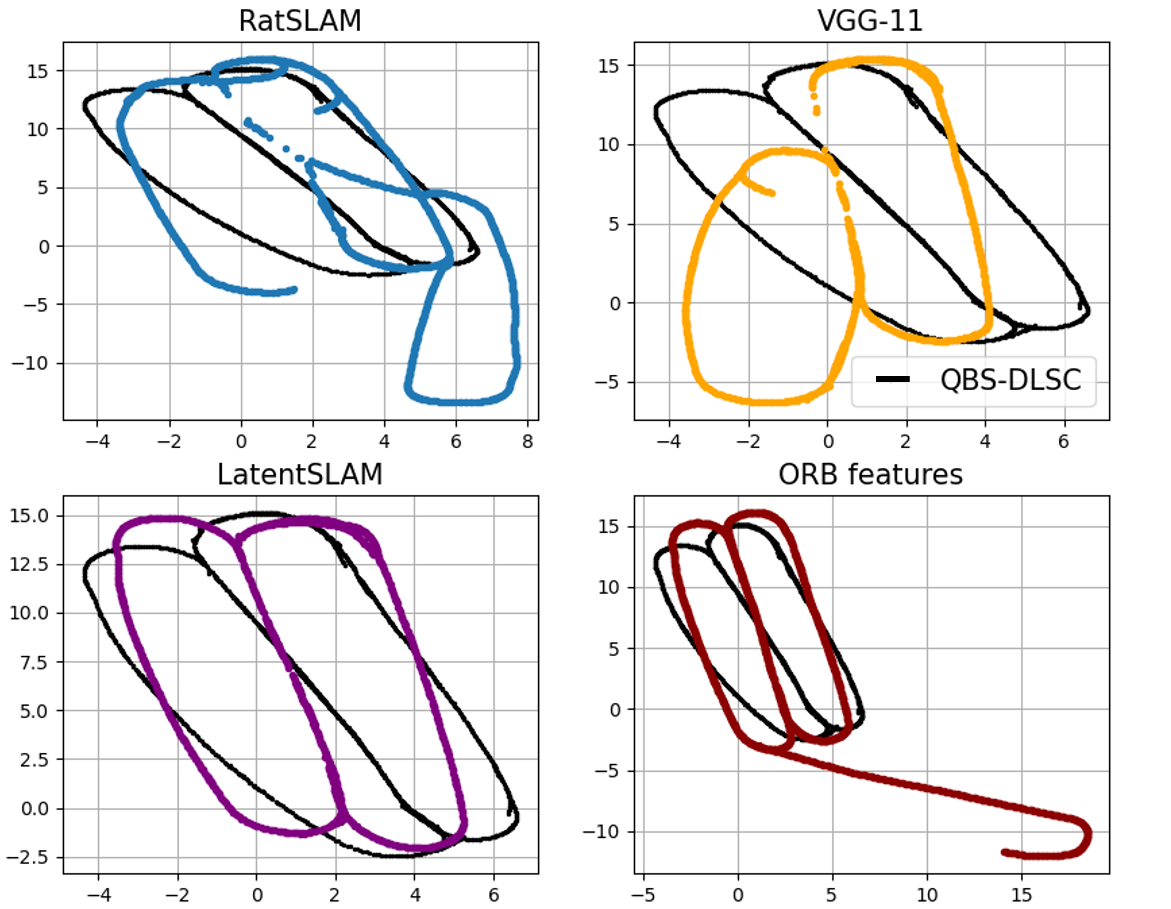}
    \caption{\textit{\textbf{Flight 3}, both red and blue paths in Fig. \ref{aliasing}.}}
    \label{seq3}
\end{figure}



Fig. \ref{seq1}, \ref{seq2} and \ref{seq3} show the SLAM trajectories obtained by each method, compared to our proposed DLSC-QBS system. As expected by prior work \cite{latentslam, alias}, the original RatSLAM system does not perform well in our ambiguous warehouse environment since its loop closure detection approach, based on the matching of raw image data, is too sensitive to the ambiguities between different views (see Fig. \ref{aliasing}). 

The pre-trained VGG-11 (deep architecture with 11 layers) used as feature descriptor performs better than the original RatSLAM for Flight 3 (see Fig. \ref{seq3}), but does not perform well overall. This confirms the observations in \cite{latentslam} where finding a good template matching threshold $\mu$ 
was shown to be hard with off-the-shelf DNNs. 

Latent-SLAM is among the top performers in Table \ref{performanceloc}, but requires the offline training of an 11-layer DNN on a dataset from the specific environment where SLAM must be performed. This makes LatentSLAM not suited for navigation in unknown environments that are not captured in datasets available beforehand, which was our goal in this work.  

The use of handcrafted ORB features, extensively used for SLAM \cite{orbslam2}, performs well in between the storage aisles (e.g., view 2 in Fig. \ref{aliasing}) but can also perform poorly 
at the end of the aisles, where lighting conditions suddenly change, leading to unreliable feature matching. Still, the SLAM back-end is able to recover the track in most cases (see Fig. \ref{seq1}, \ref{seq2}) but can also fail when, at the same time, the drift in raw odometry becomes too unreliable (see Fig. \ref{seq3}). This leads to large errors during \textit{flight 3} in Table \ref{performanceloc}.

On the other hand, our continual learning DLSC-QBS method either outperforms or either reports close MAE performance to both the pre-trained LatentSLAM and the use of ORB features, while \textit{a)} having a complexity similar to a 1-hidden-layer network; \textit{b)} not requiring any pre-training and \textit{c)} not suffering from track loss issues encountered with ORB features when lighting conditions suddenly change as the drone exits the aisles (see Fig. \ref{seq3}). Therefore, our approach might be a promising avenue for safety-critical navigation in unknown places where a dataset is not available beforehand.


\section{Conclusion}
\label{concs}
This paper has presented what is, to the best of our knowledge, one of the first continual learning SLAM systems. Our method has been experimentally validated by performing SLAM with a drone in a challenging and visually ambiguous warehouse environment, without any model pre-training, while reporting competitive performance compared to prior systems. 
We hope that this work will contribute towards safer indoor drones that can adapt to new environments on the fly.  

\section*{Acknowledgment}
We thank Prof. J. Suykens for discussing the QBS formulation of Section \ref{props}, and Dr. L. Keuninckx for his support.




\begin{thebibliography}{99}
\bibitem{uavforindoorfire} H. Surmann, D. Slomma, S. Grobelny and R. Grafe, "Deployment of Aerial Robots after a major fire of an industrial hall with hazardous substances, a report," 2021 IEEE International Symposium on Safety, Security, and Rescue Robotics (SSRR), 2021, pp. 40-47
\bibitem{ratslam} M. J. Milford, G. F. Wyeth and D. Prasser, "RatSLAM: a hippocampal model for simultaneous localization and mapping," IEEE International Conference on Robotics and Automation, 2004. 
\bibitem{orbslam2} R. Mur-Artal, J. M. M. Montiel and J. D. Tardós, "ORB-SLAM: A Versatile and Accurate Monocular SLAM System," in IEEE Transactions on Robotics, vol. 31, no. 5, pp. 1147-1163, Oct. 2015
\bibitem{contSLAM} Vödisch, N., Cattaneo, D., Burgard, W., Valada, A.. (2022). "Continual SLAM: Beyond Lifelong Simultaneous Localization and Mapping through Continual Learning."
\bibitem{latentslam} O. Çatal, W. Jansen, T. Verbelen, B. Dhoedt and J. Steckel, "LatentSLAM: unsupervised multi-sensor representation learning for localization and mapping," 2021 IEEE International Conference on Robotics and Automation (ICRA), 2021, pp. 6739-6745
\bibitem{alias} Yu, S., Wu, J., Xu, H., Sun, R., Sun, L. (2020). "Robustness Improvement of Visual Templates Matching Based on Frequency-Tuned Model in RatSLAM." Frontiers in Neurorobotics, 14.
\bibitem{nonstatdatastream} M. De Lange and T. Tuytelaars, "Continual Prototype Evolution: Learning Online from Non-Stationary Data Streams," 2021 IEEE/CVF International Conference on Computer Vision (ICCV), 2021
\bibitem{indooruwb} M. Ridolfi, N. Macoir, J. V. Gerwen, J. Rossey, J. Hoebeke and E. de Poorter, "Testbed for warehouse automation experiments using mobile AGVs and drones," IEEE INFOCOM 2019 
\bibitem{orbfeature} E. Rublee, V. Rabaud, K. Konolige, G. R. Bradski: "ORB: An efficient alternative to SIFT or SURF." ICCV 2011: 2564-2571.

\bibitem{ksvd} Mairal, J., Bach, F., Ponce, J., Sapiro, G. (2009). "Online Dictionary Learning for Sparse Coding." In Proceedings of the 26th ACM Annual International Conference on Machine Learning (pp. 689–696).

\bibitem{daubechies} Daubechies, I., Defrise, M. and De Mol, C. (2004), "An iterative thresholding algorithm for linear inverse problems with a sparsity constraint." Comm. Pure Appl. Math., 57: 1413-1457.
\bibitem{hinge} Schölkopf, B., Smola, A. (2002). "Learning with kernels : support vector machines, regularization, optimization, and beyond." MIT Press.
\bibitem{oneclass} Schölkopf, B., Williamson, R., Smola, A., Shawe-Taylor, J., Platt, J. (1999). "Support Vector Method for Novelty Detection." In Advances in Neural Information Processing Systems. MIT Press.
\bibitem{bayessurprise} Baldi, Pierre, and Laurent Itti. “Of bits and wows: A Bayesian theory of surprise with applications to attention.” Neural networks : the official journal of the International Neural Network Society (2010) 
\bibitem{stable1} H. Xu, C. Caramanis and S. Mannor, "Sparse Algorithms Are Not Stable: A No-Free-Lunch Theorem," in IEEE Transactions on Pattern Analysis and Machine Intelligence, vol. 34, no. 1, pp. 187-193, 2012
\bibitem{vgg} Simonyan, K., Zisserman, A. (2014). "Very deep convolutional networks for large-scale image recognition." 
\bibitem{lowe} Lowe, D.G. "Distinctive Image Features from Scale-Invariant Keypoints." International Journal of Computer Vision 60, 91–110 (2004). 


\end{thebibliography}
\end{document}